\documentclass[a4paper,fleqn,numbers]{cas-sc}

\usepackage[numbers]{natbib}
\usepackage[T1]{fontenc}
\usepackage{bm}
\usepackage{algorithm}
\usepackage{amsmath,amsfonts}
\usepackage{amssymb}
\usepackage{array}
\usepackage{booktabs}
\usepackage{changepage}
\usepackage{color}
\usepackage{graphicx}
\usepackage{float}
\usepackage{placeins}
\usepackage{multirow}
\usepackage{makecell}
\usepackage[most]{tcolorbox}
\usepackage{xcolor}
\definecolor{PromptGreen}{RGB}{77,158,74}
\definecolor{LightGreen}{HTML}{F2F8F5}
\definecolor{PromptPink}{HTML}{fecaba}
\definecolor{PromptBlue}{HTML}{aad8e3}
\definecolor{PromptOrange}{HTML}{ff914d}
\usepackage{listings}
\usepackage[utf8]{inputenc}
\usepackage{fvextra}
\usepackage{url}
\usepackage{listings}
\usepackage{caption}
\usepackage{xurl}
\begin{document}
\let\WriteBookmarks\relax
\def\floatpagepagefraction{1}
\def\textpagefraction{.001}
\shorttitle{MA-RAG for Query-Driven Summarization}
\shortauthors{Alamgeer et~al.}

\title [mode = title]{MA-RAG: Multi-Agent Retrieval-Augmented Generation for Query-Driven Summarization of Longitudinal Parkinson's Disease Assessments}                      

\author[1]{Sana Alamgeer}[orcid=0000-0002-6472-7570]
\ead{sanaalamgeer@gmail.com}

\author[1]{Denise Gobert}
\author[1]{Muhammad Irshad}
\author[1]{Anne H. H. Ngu}

\corref{cor1}
\cortext[cor1]{Corresponding author}

\affiliation[1]{
  organization={Texas State University},
  city={San Marcos},
  state={Texas},
  country={USA}
}

\begin{abstract}
Accurate interpretation of single-visit and longitudinal clinical assessments for Parkinson's disease is time-consuming and often depends on specialist expertise. Although large language models (LLMs) can generate natural language summaries, they frequently lack domain-specific clinical grounding and struggle to produce factually correct and temporally consistent responses for structured longitudinal assessment data. To address these limitations, we propose MA-RAG, a query-driven multi-agent retrieval-augmented generation framework that decomposes clinical reasoning into domain-specialized agents, combines structured fact extraction, and synthesizes clinically grounded summaries through a final verification stage. The framework supports four clinical analysis tasks: single-session, trajectory, comparison, and cohort summarization. We evaluate MA-RAG using objective metrics, namely Fact Precision, Hallucination Rate, Temporal Fidelity, and Semantic Similarity, together with subjective evaluations conducted by clinical experts. Compared to Traditional, RAG-only, and Single-agent RAG baselines, MA-RAG substantially improves factual correctness, achieving up to a 122\% relative increase in Fact Precision (from 0.436 to 0.990) and reducing the Hallucination Rate by up to 98\% (from 0.564 to 0.010), while consistently receiving top ratings from clinical experts for organization and clinical usefulness. These results demonstrate that domain-specialized multi-agent reasoning enables reliable query-driven summarization of structured longitudinal clinical assessment data.
\end{abstract}




\maketitle

\section{Introduction}\label{sect:introduction}

\begin{figure}
 \centering
 \includegraphics[width=0.99\linewidth]{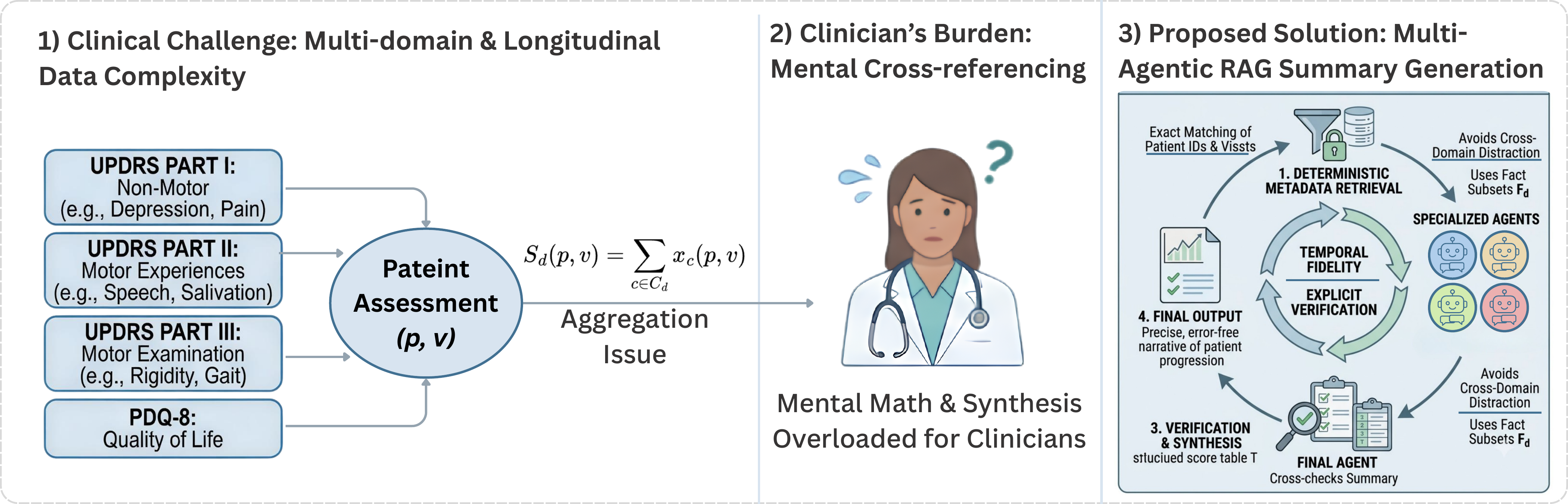}
 \caption{Overview of 1) the clinical challenge, 2) clinician burden, and 3) the MA‑RAG automated interpretation/summary generation system for multi‑domain longitudinal Parkinson’s assessments.}
 \label{fig:intro}
\end{figure}

Parkinson's disease (PD) is a progressive neurodegenerative disorder characterized by motor and non-motor symptoms that require continuous multimodal assessment. The Unified Parkinson's Disease Rating Scale (UPDRS)~\cite{Goetz2007MDSUPDRS} serves as the gold standard for clinical evaluation, comprising three primary domains used in routine practice: Part I) Evaluates non-motor experiences of daily living, Part II) Assesses motor aspects of activities of daily living (ADLs), and Part III) Measures motor function through clinician-administered examination. Each domain consists of multiple items scored on an ordinal scale from $0$ to $4$, where higher values indicate greater symptom burden. In clinical practice, the Parkinson's Disease Questionnaire-8 (PDQ-8) is commonly used alongside the UPDRS to complement motor and non-motor assessments with a patient-reported measure of health-related quality of life (QoL). The PDQ-8 is a validated short-form version of the PDQ-39, comprising eight items that capture quality of life across multiple dimensions~\cite{Peto1995PDQ39}. For a given patient $p$ at visit $v$, let $x_c(p,v)$ denote the raw score of item $c \in C_d$ within domain $d$. The aggregate domain score $S_d(p,v)$ is computed as the summation of its constituent items,
\begin{equation}
    S_d(p,v) = \sum_{c \in C_d} x_c(p,v),
\end{equation}

where $d \in \{\text{Non-Motor}, \text{ADL}, \text{Motor}, \text{QoL}\}$ corresponds to UPDRS Parts I, II, III, and the PDQ-8, respectively. As illustrated in Figure~\ref{fig:intro}, this mathematical structure captures the multidimensional nature of PD progression but introduces significant complexity for clinicians, where they face substantial challenges in synthesizing these structured data into actionable longitudinal insights. First, documenting assessments and manually calculating clinical scores is time-consuming, limiting the practicality of comprehensive evaluation during routine clinical encounters. Second, although complementary assessment domains are routinely collected, they remain largely isolated from one another, forcing clinicians to mentally integrate single-domain $d$ findings when forming an overall clinical impression. Third, interpreting longitudinal disease progression is difficult because clinicians must manually compare assessment results across multiple visits and relate changes in individual domain scores, without a unified view of how the patient's condition evolves over time. As a result, clinicians often rely on static staging scales, such as the Hoehn and Yahr scale~\cite{Hoehn1967}, which may not fully reflect short-term changes in a patient's condition.

Existing computational tools, such as large language models (LLMs)~\cite{singhal2022llm, chen2023meditron70b, 9222332, Gridach2026}, and retrieval-augmented generation (RAG)~\cite{lewis2021rag, gao2024rag}, remain insufficient for reliable clinical decision support: 1) standard LLMs frequently hallucinate numeric values when processing multi-domain inputs and fabricate scores absent from source documentation, 2) models exhibit selective attention bias by focusing disproportionately on prominent motor features while neglecting critical non-motor or QoL indicators embedded in the same record, 3) RAG systems without structured reasoning pipelines fail to maintain temporal fidelity across longitudinal visits and produce summaries that violate chronological order or misattribute baseline changes, and 4) current approaches lack explicit verification mechanisms against underlying fact tables, allowing factual errors to propagate unchecked into final clinical narratives. Recent studies further demonstrate that relying solely on autonomous LLMs or agentic reasoning, without rule-based verification against trusted evidence, can amplify hallucinations and reasoning errors, potentially leading to harmful downstream decisions in high-stakes clinical workflows~\cite{Huang_2025}.

We propose a multi-agent retrieval-augmented generation framework (MA-RAG) designed specifically for query-driven longitudinal summarization of multimodal clinical assessments of patients with Parkinson's disease. Our approach addresses the aforementioned limitations through three core innovations:
\begin{enumerate}
    \item We employ metadata-based retrieval rather than embedding similarity search, ensuring exact matching of patient identifiers and visit timestamps to eliminate retrieval ambiguity. 
    \item We decompose clinical reasoning into domain-specialized agents that operate exclusively on precomputed fact subsets $F_d$, preventing cross-domain distraction and ensuring complete coverage of all relevant findings. 
    \item We implement a dedicated verification stage in which a Final Agent cross-checks upstream narratives against a structured score table $T$ before synthesis, ensuring that every reported value is grounded in the extracted facts, enabling accurate longitudinal reasoning over heterogeneous clinical assessments. 
    \item We validate MA-RAG through a comprehensive evaluation that combines objective metrics, including Fact Precision, Hallucination Rate, Temporal Fidelity, and Semantic Similarity, with subjective evaluations conducted by the clinical experts.
\end{enumerate}

The remainder of this paper is organized as follows. Section~\ref{sect:related_work} reviews existing approaches to clinical summarization and identifies gaps in handling multimodal longitudinal data. Section~\ref{sect:methodology} details the MA-RAG architecture, and Section~\ref{sect:impl_details} describes the datasets, knowledge base construction, and experimental setup. Section~\ref{sect:results_discuss} presents the evaluation results alongside computational cost analysis, ethical considerations, and limitations. Finally, Section~\ref{sect:conclusion} summarizes key contributions and outlines future directions.

\section{Related Work}\label{sect:related_work}
The evolution of clinical summarization has progressed through four distinct technical paradigms, each addressing specific limitations of its predecessor while introducing new failure modes in longitudinal multimodal reasoning. Early approaches relied on template-based natural language generation (NLG) systems that mapped structured assessment scores to fixed textual patterns~\cite{jamia1997, Hunter2011BTNurse, HUNTER2012157}. While these systems guaranteed factual consistency by construction, their rigid output templates lacked the semantic flexibility required to synthesize cross-domain relationships or adapt to variable visit counts~\cite{Tang2023, lyu2025nlp, CARENZO2026}. For example, a patient exhibiting stable motor scores but rapidly worsening depression and sleep disturbance would receive an identical narrative template to a patient with uniform progression across all domains, obscuring clinically critical discordance patterns that require dynamic synthesis.

Abstractive transformer models~\cite{11437317, 11102242} overcame the rigidity of template-based methods by learning latent representations of clinical narratives from large corpora, enabling fluent adaptation to diverse trajectory patterns. Models such as ClinicalBERT~\cite{huang2020} and BioGPT~\cite{Alleva2025} demonstrated superior domain adaptation compared to rule-based methods. However, these autoregressive generators operate as black-box probability distributions without explicit grounding mechanisms, causing them to hallucinate numeric values when processing structured multimodal inputs~\cite{Singhal2023}. In our own empirical evaluations, such models achieve only 43.6\% fact precision on longitudinal queries because they conflate similar-looking score patterns across visits and fabricate plausible but nonexistent baseline differences/deltas. A concrete failure case involves a model reporting ``Motor score improved from 28 at Visit 1 to 22 at Visit 3'' when the actual records show 28 at Visit 1, 31 at Visit 2, and 35 at Visit 3, i.e., the model invents both the intermediate value and the direction of change based on probabilistic token prediction rather than retrieved evidence.

Retrieval-augmented generation (RAG)~\cite{lewis2021rag, gao2024rag, ai6090226} was introduced to mitigate hallucination by conditioning generation on retrieved evidence chunks rather than parametric memory alone. Standard RAG pipelines using dense vector similarity search improved factual grounding for single-session summaries by retrieving semantically relevant passages~\cite{gao2024rag}, directly resolving the fabrication failures observed in vanilla LLMs by anchoring outputs to verifiable text spans. Yet this approach fails catastrophically for longitudinal clinical assessments because embedding similarity cannot enforce exact patient-visit alignment. When querying ``trajectory of patient 3'', cosine-similarity retrieval returns top-k documents from multiple patients whose text embeddings are semantically close but belong to different individuals, violating the fundamental requirement of patient-specific temporal coherence~\cite{asai2023}. Furthermore, RAG retrieves unstructured text fragments rather than structured fact tables, forcing the LLM to perform arithmetic aggregation and baseline delta computation during generation, a task at which autoregressive models consistently fail due to their lack of symbolic reasoning capabilities~\cite{liu2025maths}.

Agentic frameworks emerged to decompose complex clinical reasoning into specialized subtasks coordinated through the usage of tools and multi-step planning~\cite{XU2026105045}. Systems like MedAgent~\cite{tang2024} and ClinicalAgents~\cite{Liu2026} route queries to domain-specific modules and invoke external calculators for score aggregation, directly resolving the arithmetic failure mode of standard RAG by offloading numerical operations to deterministic tools. While this architecture reduces cross-domain distraction by isolating domain-specific reasoning into separate agents, existing implementations lack explicit verification stages. Generated narratives pass directly to output without cross-checking against underlying structured facts, allowing subtle errors to propagate undetected into final summaries. A concrete failure case involves an agentic system correctly computing individual domain scores via external tools but then generating a narrative stating ``ADL worsened by 5 points from baseline'' when the tool output actually showed a 3-point improvement. Additionally, current agentic systems rely on embedding-based retrieval for knowledge access, inheriting the same patient-mixing failures that plague standard RAG when processing longitudinal records with high inter-patient semantic similarity.

Despite these advances, no existing framework combines metadata-based retrieval, domain-specialized agentic decomposition, and explicit fact-table verification in a unified pipeline for longitudinal multimodal clinical assessment summarization. Moreover, existing evaluation practices remain limited. Widely used metrics, including ROUGE~\cite{Lin2004ROUGEAP}, BLEU~\cite{Papineni2002BleuAM}, BERTScore~\cite{zhang2020sbert}, and SBERT similarity~\cite{reimers2019sbert}, fail to adequately capture factual correctness and temporal consistency~\cite{Vasilev2025, Asgari2025}. As a result, they can assign high scores to summaries that match the reference text lexically or semantically while misreporting visit ordering, baseline anchoring, or numeric clinical findings. Together, these architectural and evaluation limitations motivate the development of more reliable frameworks for summarization of longitudinal multimodal clinical assessments.

\section{Methodology}\label{sect:methodology}
\begin{figure}
  \centering
  \includegraphics[width=\linewidth]{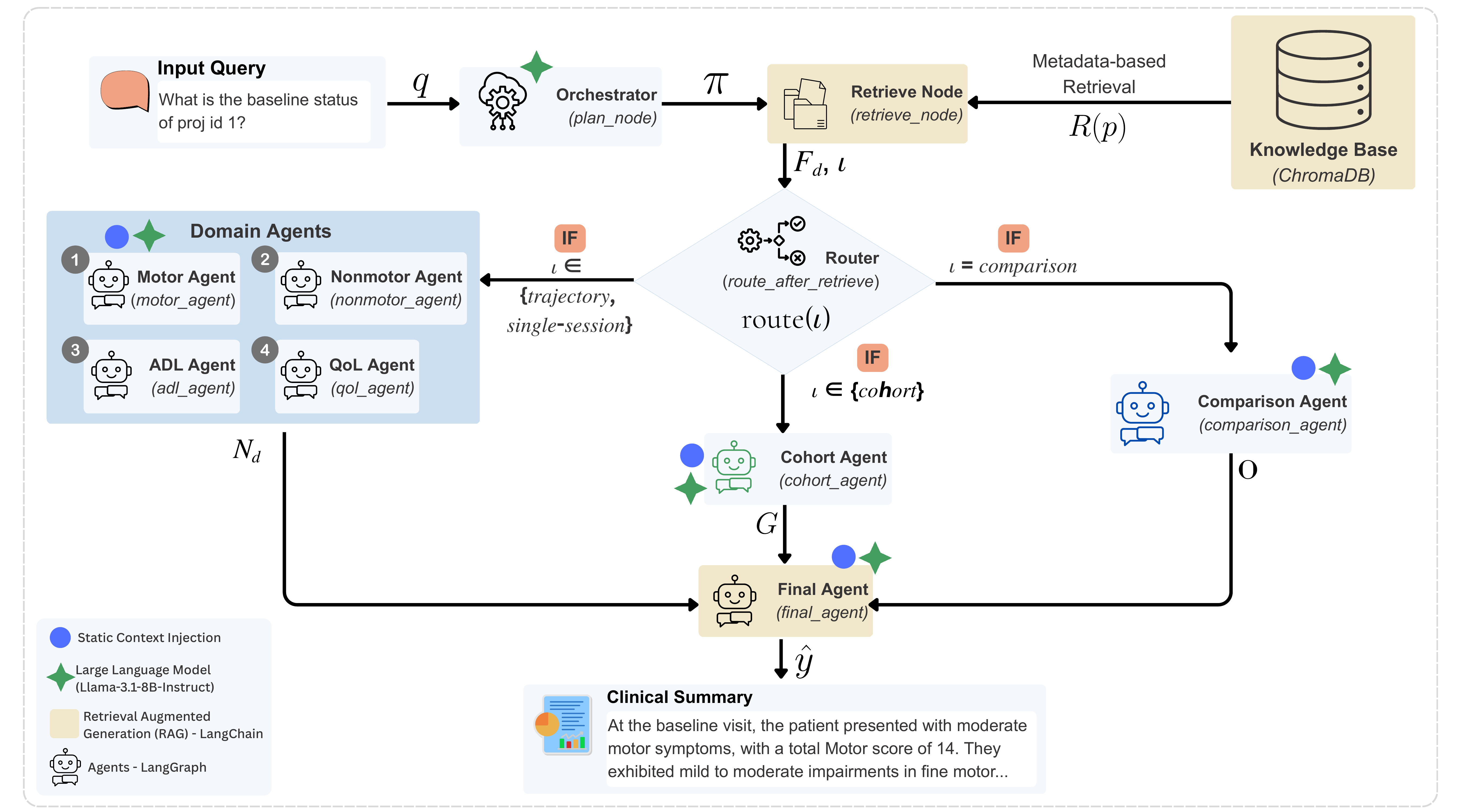}
  \caption{Overview of the MA-RAG architecture: Given a query $q$, the Orchestrator plans retrieval, the Retrieve Node fetches patient records from the ChromaDB Knowledge Base via metadata-based matching $R(p)$, and a router directs the flow to Domain, Cohort, or Comparison Agents. The Final Agent synthesizes these outputs into a clinical summary $\hat{y}$.}
  \label{fig:flowchart}
\end{figure}

The proposed framework, MA-RAG, generates clinical summaries in response to a query $q$ over longitudinal patient records. The architecture, shown in Figure~\ref{fig:flowchart}, consists of three components. First, the RAG pipeline comprises a Knowledge Base, a Retrieve Node, and a Final Agent. The Knowledge Base (KB) stores patient session records together with associated metadata (details are provided in section~\ref{sect:impl_details}). The Retrieve Node retrieves evidence by metadata-based matching, and the Final Agent grounds its synthesis in the retrieved evidence before generating the final summary. 
Although the current implementation indexes structured assessment documents, the document-oriented representation naturally supports future integration of additional clinical evidence, such as clinician-authored narratives, through the same vector database and RAG retrieval pipeline. 
Second, query interpretation is performed by a set of LLM-driven agents. Third, the outputs of these agents are synthesized to produce the final clinical summary.

\subsection{Task Definition}

Let $\mathcal{P}$ denote the set of patients and $\mathcal{V}_p$ the ordered set of visits for patient $p \in \mathcal{P}$. Each visit contains one or more clinical assessments, represented as domain-specific scores and item-level responses from a set of instruments (e.g., UPDRS and PDQ-8). Given a natural language query $q$, MA-RAG retrieves relevant patient records and reasons over the requested clinical assessments to generate a clinically grounded summary $\hat{y}$, which preserves the requested temporal context, integrates findings across the selected assessment domains, and maintains factual correctness.

MA-RAG supports four analysis types, each corresponding to a distinct clinical summarization task: (1) \textbf{Single-session analysis}: summarizes one clinical encounter for a single patient, typically a baseline or latest visit (e.g., What is the baseline status of proj id 5 using UPDRS?''); (2) \textbf{Trajectory analysis}: summarizes longitudinal disease progression by comparing clinical assessments across multiple visits of a single patient (e.g., Show me the UPDRS trajectory of proj id 5.''); (3) \textbf{Comparison analysis}: contrasts disease progression between multiple patients over one or more visits (e.g., Compare the disease progression on UPDRS between proj id 3 and proj id 8.''); and (4) \textbf{Cohort analysis}: performs population-level analyses to identify score extremes, progression patterns, and patient risk profiles across the cohort (e.g., Which patient has the highest UPDRS score?'').

\subsection{Orchestrator}

The Orchestrator is the first stage of the pipeline. It receives the user's query $q$ written in natural language and converts it into a structured, machine-readable plan $\pi$ that tells the rest of the system \emph{what} to analyze, \emph{for whom}, and \emph{how}.

The Orchestrator (\texttt{plan\_node}) is driven by a call to an LLM $\mathcal{L}_{plan}$, where the model is given $q$ together with a system prompt (see Appendix~\ref{app:system_prompts}), and is asked to parse $q$ into four fields, $\pi = \mathcal{L}_{plan}(q)$:

\begin{itemize}
    \item $P$ is the set of \textbf{patient identifiers} mentioned in $q$ (e.g., $P = \{1\}$). If $q$ refers to all patients, or if no identifier can be recovered, $P$ defaults to the full set of patient identifiers available in the KB.
    \item $a$ is the \textbf{analysis type}, one of $\{trajectory,\ comparison,\ cohort,\ single\text{-}session\}$. This is selected by matching keywords in $q$ against rules given to the model (e.g., comparison for ``compare'', and single-session for ``baseline'' or ``latest'').
    \item $M$ is the \textbf{instruments} of interest, a subset of $\{updrs,\ pdq8\}$, indicating which clinical scale(s) the query concerns.
    \item $\phi$ is the \textbf{visit filter}, indicating which visits are relevant (e.g., $none$, $baseline$, $latest\text{-}visit$, or $first\text{-}N\text{-}visits$). When $q$ specifies a count of visits (e.g., ``the first 2 visits''), the integer $N$ is extracted directly from $q$ and substituted into $\phi$ (e.g., $\phi = first\text{-}2\text{-}visits$).
\end{itemize}

Together, these four fields form the plan $\pi = (P,\ a,\ M,\ \phi)$. For example, for the query $q =$ ``compare the latest visit of patients 1 and 4,'' the Orchestrator returns: $P = \{1, 4\}$, $a = comparison$, $M = \{updrs, pdq8\}$, and $\phi = latest\_visit$.

If the language model's output cannot be parsed into all four fields (for example, if it omits a field or uses unexpected formatting), the Orchestrator substitutes a safe default plan: $a = trajectory$, $M = \{updrs, pdq8\}$, and $\phi = none$, ensuring the pipeline can still proceed rather than failing outright. The output of this stage $\pi$ is passed forward to the next stage. Critically, the participant set $P$ constrains all subsequent stages, ensuring that retrieval and downstream reasoning operate exclusively on data from patients in $P$.

\subsection{Retrieve Node}
The input to this stage is the original query $q$ and the plan $\pi = (P, a, M, \phi)$ from the Orchestrator. The job of Retrieve Node (\texttt{retrieve\_node}) is to gather and organize the exact patient data needed to answer the query.

The Retrieve Node first maps the analysis type $a$ already present in $\pi$ directly onto an internal intent label $\iota$,
\begin{equation}
    \iota = map(a), \qquad a \in \{trajectory,\ comparison,\ cohort,\ single\text{-}session\},
\end{equation}
where $map(\cdot)$ is a fixed lookup table that reconciles minor naming differences between the Orchestrator's output and the pipeline's internal labels (e.g., $single\text{-}session$ is mapped to $single\_session$). 

With the intent resolved, the Retrieve Node proceeds through a sequential pipeline to assemble the fact table ($F_d$). First, for every patient identifier $p \in P$, the corresponding session records are fetched from the KB by exact metadata match, $R(p) = \{ X(p,v) \mid \text{proj\_id} = p,\ \text{granularity} = session \}$, where $X(p,v)$ denotes the stored document for patient $p$ at visit $v$. Second, from $R(p)$, the domain scores $S_d(p,v)$ and their item-level scores are extracted for the domains $d$ relevant to the query, producing a structured fact table $F$ indexed by patient, visit, and domain. Finally, the resulting table is split into four domain-specific subsets, $F_{motor}$, $F_{adl}$, $F_{nonmotor}$, and $F_{qol}$, so that each downstream domain agent receives only the facts relevant to its own specialty. 

As a working example, let $P = \{1\}$, $a = single\text{-}session$, and $\phi = baseline$, the Retrieve Node maps $\iota = single\text{-}session$ directly from $a$, fetches all session records for patient 1, and extracts the visit-1 motor score $S_{motor}(1,1) = 14$, representing that patient $1$ has a total Motor score of $14$ at visit $1$. The output of this stage is the intent $\iota$ together with the fact table $F$ and its domain subsets $F_{motor}, F_{adl}, F_{nonmotor}, F_{qol}$, all passed to the Router.

\subsection{Router}
The Router (\texttt{route\_after\_retrieve}) is a rule-based function that selects the next stage of the pipeline based solely on the intent $\iota$ from the previous stage, without invoking a language model:

\begin{equation}
    route(\iota) =
    \begin{cases}
    \text{Comparison Agent}, & \iota = comparison \\
    \text{Cohort Agent}, & \iota \in \{cohort,\ risk\} \\
    \text{Domain Agents}, & \iota \in \{trajectory,\ single\text{-}session\}
    \end{cases}
\end{equation}

When $\iota = comparison$, the Comparison Agent is invoked with the full fact table $F$, since comparing patients requires access to all domains simultaneously. When $\iota \in \{trajectory, single\_session\}$, the Domain Agents are invoked, each receiving only its corresponding subset, $F_{motor}$, $F_{adl}$, $F_{nonmotor}$, or $F_{qol}$, so that it acts solely on the data relevant to its specialty. The output of this stage is the identity of the invoked agent, together with the data passed to it.

\subsection{Domain Agents}

When the path of the Domain Agents is selected, four agents are invoked in sequence: the Motor Agent (\texttt{motor\_agent}), the ADL Agent (\texttt{adl\_agent}), the Non-Motor Agent (\texttt{nonmotor\_agent}), and the QoL Agent (\texttt{qol\_agent}). Each agent is given its own fact subset $F_d$, the query $q$, and a static reference text $C_d$ (see Appendix) describing that domain's scale, and calls an LLM $\mathcal{L}_{domain}$ to write a progress narrative, $N_d = \mathcal{L}_{domain}(q, F_d, C_d)$. If $F_d$ is empty, the agent simply skips and returns no narrative.

Recall that $S_d(p,v)$ denotes the score of domain $d$ for patient $p$ at visit $v$, as recorded in $F_d$. Each agent is instructed to compute, for every visit $v > 1$, the change from the previous visit, $\Delta S_d(p,v) = S_d(p,v) - S_d(p,v-1)$, and the cumulative change from baseline,
\begin{equation}
    \Delta S_d^{base}(p,v) = S_d(p,v) - S_d(p,1),
\end{equation}
labeling the direction as improvement when the score decreases and worsening when it increases. The output of this stage is a progression summary for each domain, $N_{motor}$, $N_{adl}$, $N_{nonmotor}$, and $N_{qol}$, passed to the Final Agent.

\subsection{Comparison Agent}
When the Comparison Agent (\texttt{comparison\_agent}) is invoked, it receives the fact table $F$ and the query $q$ from the Retrieve Node. Then, it first builds, through a rule-based logic, a table $B$ that lists the score of every domain at every visit for every patient in $P$, along with whether each score improved, worsened, or stayed the same compared to the previous visit. For example, a single row of $B$ might read: \emph{Patient 1, Visit 2, Motor $=25$, worsened ($+5$)}, indicating that patient 1's Motor score increased by 5 points from Visit 1 to Visit 2. This table $B$, the query $q$, and the combined reference text $C = \bigcup_d C_d$ describing all four domain scales are then passed to an LLM call $\mathcal{L}_{comparison}$, which writes a summary comparing the patients across the four domains, $\mathcal{O} = B \oplus \mathcal{L}_{comparison}(q, B, C)$, where $\oplus$ denotes text concatenation. The output $\mathcal{O}$ is then passed to the Final Agent.

\subsection{Cohort Agent}

When the Cohort Agent is invoked (\texttt{cohort\_agent}), it receives the fact table $F$ and the query $q$ from the Retrieve Node. Unlike the Domain Agents and the Comparison Agent, which narrate an individual patient's or a small group's trajectory, the Cohort Agent aggregates statistics across the entire patient set $P$ to answer population-level questions, such as identifying extremes or averages.

The agent first computes, for every patient $p \in P$ and every domain $d$, the slope of that patient's progression across visits,
\begin{equation}
slope_d(p) = \frac{S_d(p, v_{last}) - S_d(p, v_{first})}{v_{last} - v_{first}},
\end{equation}
together with the maximum, minimum, and average of $S_d$ across all patients. From these values, a specific statistic $\mu$ of interest, for example the patient with the highest slope in a given domain, is selected by matching keywords in $q$ against a fixed rule set (e.g., ``highest'' or ``worst'' maps to maximum total burden, ``fastest'' combined with a particular domain name maps to that domain's steepest slope, and ``stable'' maps to the smallest absolute slope). The statistic $\mu$, the query $q$, and the combined reference text $C = \bigcup_d C_d$ describing all four domain scales are then passed to an LLM call $\mathcal{L}_{cohort}$, which writes a summary explaining the cohort finding in clinical terms, $G = \mathcal{L}_{cohort}(q, \mu, C)$. The output $G$ is passed to the Final Agent.

\subsection{Final Agent}

The Final Agent (\texttt{final\_agent}) receives the output produced by whichever branch was invoked in the prior stage, either the Domain Agents ($N_{motor}, N_{adl}, N_{nonmotor}, N_{qol}$), the Comparison Agent ($\mathcal{C}$), or the Cohort Agent ($G$), together with the fact table $F$. The role of the agent is not simply to write a summary, but to check the prior narrative against the underlying facts before doing so, and only then produce the final clinical summary. The agent first computes, from $F$, a score table $T = build(F)$, listing $S_d(p,v)$ for every domain $d$ and visit $v$, annotating any visit where one domain's score is zero while another is elevated for the same patient as a potential incongruence requiring closer verification.

A dispatcher then checks which prior agent produced output and selects the matching synthesis mode $m \in \{\text{cohort}, \text{comparison}, \text{single-session}, \text{trajectory}\}$,
\begin{equation}
    m =
    \begin{cases}
    \text{cohort}, & G \neq \emptyset \\
    \text{comparison}, & G = \emptyset \ \wedge\ \mathcal{C} \neq \emptyset \\
    \text{single-session}, & G = \emptyset \ \wedge\ \mathcal{C} = \emptyset \ \wedge\ a = single\text{-}session \\
    \text{trajectory}, & \text{otherwise},
    \end{cases}
\end{equation}
where $\wedge$ denotes the logical AND operator, requiring all conditions on a given line to hold simultaneously for that case to apply, and $a$ is the analysis type carried in the plan $\pi$. In each mode $m$, an LLM call $\mathcal{L}_m$ is instructed to verify the relevant upstream narrative $r$ against $T$ and $F$ before writing the summary, so that only confirmed facts are included,
\begin{equation}
    \hat{y} = \mathcal{L}_m(q, r, T, F),
\end{equation}
where $r$ denotes $G$, $\mathcal{C}$, or $N_{motor}, N_{adl}, N_{nonmotor}, N_{qol}$, depending on $m$.

For example, in trajectory mode, the model writes one paragraph on progression over all visits anchored to the baseline scores $S_d(p,1)$, using the cumulative change $\Delta S_d^{base}(p,v)$ defined earlier, followed by a paragraph on cross-domain patterns and one on domains needing closest monitoring. In comparison mode, the model instead writes one paragraph per domain, directly contrasting the patients using $B$, followed by a paragraph on cross-patient discordance and divergence points. The output of this final stage is the final clinical summary $\hat{y}$, returned to the user in response to $q$.

\section{Implementation Details}\label{sect:impl_details}
\subsection{Datasets}
We use a longitudinal patient dataset collected under Texas State University IRB protocol \#6785. The dataset comprises two complementary clinical assessment modalities: the Unified Parkinson's Disease Rating Scale (UPDRS) and the Parkinson's Disease Questionnaire-8 (PDQ-8). The dataset comprises 44 adults with Parkinson's disease (9 females and 35 males), diagnosed at Hoehn and Yahr stages 1--3 at baseline, with a mean age of 67.1 years and a mean disease duration of 4.3 years. Each participant completed four assessment sessions over 24 weeks. The UPDRS records clinician-assessed motor, ADL, and non-motor symptoms, whereas the PDQ-8 provides a patient-reported measure of quality of life (QoL).

The clinical records are stored as session-level CSV files, where each row corresponds to a patient visit. Each record contains a patient identifier $p$, a raw visit timestamp $t_{\mathrm{raw}}$, and the item-level assessment scores. During preprocessing, the timestamp is converted into a standardized date $t$, and the visits of each patient are sorted chronologically. A visit index $v \in \{1,\ldots,N_p\}$ is then assigned, where $v=1$ denotes the baseline visit and $N_p$ is the total number of visits for patient $p$. Instead of organizing the records by assessment instrument (UPDRS and PDQ-8), we organize them into four clinical domains, motor, ADL, non-motor, and QoL, reflecting how clinicians typically assess patient function.

\subsection{Knowledge Base Construction}
The domain scores are converted into a short text document for each visit. Given a visit $(p,v)$ and its scores $S_{motor}$, $S_{adl}$, $S_{nonmotor}$, and $S_{pdq8}$, the document $D(p,v)$ states the patient identifier $p$, the visit index $v$, the visit date $t$, and each domain score together with its item-level scores. For example, $D(p,v)$ may read: \textit{Patient 12, Visit 2, 2021-03-15, Motor=14 (Tremor=2, Gait=3, etc.), ADL=8 (Dressing=2, Walking=2, etc.), NonMotor=5 (Sleep=2, Fatigue=1, etc.), PDQ8=11 (Mobility=2, Concentration=1, etc.)}. Each document is labeled with two metadata fields: \textit{instrument}, indicating which scale it belongs to (UPDRS or PDQ-8), and \textit{granularity}, indicating whether it is a session (single-visit) or a longitudinal (multi-visit) patient summary.

Each document $D(p,v)$ is then encoded into a dense vector by a sentence embedding model $E$,
\begin{equation}
    \mathbf{e}(p,v) = E\big(D(p,v)\big), \quad \mathbf{e}(p,v) \in \mathbb{R}^{d}
\end{equation}

where $d$ is the embedding dimension of $E$. All documents, their embeddings, and their metadata are stored in a persistent ChromaDB collection, denoted the Knowledge Base. 
At inference time, MA-RAG retrieves single-visit documents from the KB through exact match on their metadata fields. For example, given the query \textit{``What is the baseline status of proj id 1?''}, MA-RAG retrieves all documents $D(1,v)$ for patient $p=1$ with $\text{granularity}=\text{session}$, and then applies the visit filter \textit{baseline} to select only the document corresponding to $v=1$.

\subsection{Static Clinical Context}

Static domain contexts $C_d$ are constructed for $d \in \{\text{motor}, \text{ADL}, \text{nonmotor}, \text{QoL}\}$, where each $C_d$ encodes the clinical definitions of the assessment items, item-level scoring guidelines, clinical interpretation, and reliability statistics relevant to the corresponding domain, extracted from UPDRS~\cite{Goetz2007MDSUPDRS} and PDQ-39~\cite{Peto1995PDQ39}. The combined static context, provided to the Cohort, Comparison, and Final agents, is defined as
\begin{equation}
C = \bigcup_{d} C_d.
\end{equation}

Unlike patient-specific records, $C$ is not stored in or retrieved from the KB. First, this knowledge is invariant across queries, making repeated retrieval unnecessary. Second, retrieval-based context is subject to token-budget constraints and truncation, which may omit clinically important information. Instead, each domain-specific agent is initialized with its corresponding static context $C_d$, ensuring complete and consistent clinical knowledge during inference (see Appendix~\ref{app:static_contexts} for details).

\subsection{Experimental Setup}

We employ ChromaDB as the vector store for the retrieval and KB components, computing document embeddings using the \texttt{all-MiniLM-L6-v2} model from the \texttt{sentence-transformers} library. We use \texttt{Llama-3.1-8B-Instruct} as the underlying LLM across all agent nodes via HuggingFace's \texttt{transformers} library. For deterministic generation and reproducibility, LLM calls are configured with greedy decoding, a maximum output length of 1,200 tokens, and a repetition penalty of 1.1. We orchestrate the multi-agent workflow using LangGraph, with node and edge definitions built on LangChain message primitives, and expose the system through a lightweight Gradio interface that serves as a prototype interaction layer, hosted on a single NVIDIA A100 GPU (80 GB VRAM).

\subsection{Evaluation Protocol}\label{subsect:eval_proto}

For the offline evaluation protocol, we construct gold summaries for single-session and longitudinal queries using GPT-5.5~\cite{openai2026gpt55} with clinician assistance, and treat them as ground truth. We compare generated summaries against these gold summaries using four metrics: Fact precision~\cite{Asgari2025}, hallucination rate~\cite{PhysioNet2025}, temporal fidelity~\cite{kruse-etal-2025-large}, and SBERT similarity~\cite{reimers2019sbert}.

Fact precision is defined as the fraction of numeric tokens $\mathcal{N}_{model}$ in a summary that also appear in the extracted fact table $F$. Here, $\mathcal{N}_{model}$ denotes the set of numeric tokens extracted from a generated summary, and $\mathcal{N}_{facts}$ denotes the set of numeric tokens extracted from $F$,

\begin{equation}
    \text{FactPrecision} = \frac{|\mathcal{N}_{model} \cap \mathcal{N}_{facts}|}{|\mathcal{N}_{model}|}
\end{equation}

The hallucination rate, in contrast, is the fraction of $\mathcal{N}_{model}$ absent from $\mathcal{N}_{facts}$,

\begin{equation}
    \text{HallucinationRate} = \frac{|\mathcal{N}_{model} \setminus \mathcal{N}_{facts}|}{|\mathcal{N}_{model}|}
\end{equation}

Temporal fidelity checks whether the visit indices $(v_1, \dots, v_k)$ mentioned in a summary appear in increasing order, where $v_i \in \mathbb{N}$ denotes the $i$-th visit index extracted from the summary, and $k$ is the total number of visit indices mentioned,

\begin{equation}
    \text{TemporalFidelity} = \mathbb{1}\big[(v_1, \dots, v_k) = \text{sort}(v_1, \dots, v_k)\big]
\end{equation}

Lastly, SBERT similarity is the cosine similarity between sentence embeddings, under \texttt{all-MiniLM-L6-v2}, of a generated summary and its gold summary. Here, $\mathbf{e}_{model} \in \mathbb{R}^d$ denotes the sentence embedding of the generated summary, and $\mathbf{e}_{gold} \in \mathbb{R}^d$ denotes the sentence embedding of the gold summary, both of dimension $d$,

\begin{equation}
    \text{SBERTSim} = \frac{\mathbf{e}_{model} \cdot \mathbf{e}_{gold}}{\lVert \mathbf{e}_{model} \rVert \, \lVert \mathbf{e}_{gold} \rVert}
\end{equation}

We compute all four metrics per query, and report them as a mean, grouped by task type (single-session and longitudinal).

\section{Results and Discussion}\label{sect:results_discuss}

\subsection{Offline Evaluation}
\subsubsection{Comparison Across Different Frameworks}
To assess MA-RAG's factual and temporal reliability in a controlled setting, we conducted two offline experiments, one targeting single-session queries and the other targeting longitudinal queries. For the single-session experiment, we queried the system with \textit{``What are the \{instrument\} scores and findings for \{proj id\} on \{visit date\}?''}, iterating over all 44 participants and every visit date recorded for each participant. For the longitudinal experiment, we queried the system with \textit{``What is the trajectory of \{instrument\} for \{proj id\}?''}, iterating over all 44 participants across their full visit history. In both experiments, we substituted \{instrument\} with UPDRS alone, then with UPDRS and PDQ-8 combined, yielding four evaluation conditions in total. We compared the Single-session outputs against the corresponding single visit record, while longitudinal outputs were compared against gold longitudinal summaries, prepared beforehand (as described in Section~\ref{subsect:eval_proto}).

We evaluated both UPDRS alone and UPDRS combined with PDQ-8 because these instruments are routinely interpreted together in clinical practice to assess disease severity and quality of life. The combined setting presents a more challenging retrieval task, requiring MA-RAG to retrieve and reason over a larger, heterogeneous set of clinical facts while preserving domain separation and temporal consistency. This evaluation assesses the framework's ability to maintain factual correctness and temporal coherence as the complexity of the retrieved evidence increases.

We compared MA-RAG against three baseline frameworks to evaluate the contributions of structured retrieval, fact extraction, multi-agent domain specialization, and final verification. All baseline methods had access to the full static domain context set $C = \bigcup_d C_d$, utilized the same ChromaDB KB, and ran on the same underlying LLM.
\begin{enumerate}
    \item \textbf{Traditional Baseline:} Provides the LLM directly with the natural language query $q$, the static domain context $C$, and the complete set of unindexed patient records serialized as a structured JSON/dictionary object within the prompt, bypassing both retrieval and agent decomposition. This method tested whether directly supplying full, unindexed patient records within the context window is sufficient without retrieval filtering or agentic workflow.
    \item \textbf{RAG-only Baseline:} Retrieves relevant patient records from the KB using metadata matching $R(p)$. The raw text documents $X(p,v)$ are passed directly as unstructured passages into the prompt alongside $q$ and $C$, requiring a single LLM call to perform retrieval-grounded reasoning without explicit fact extraction ($F$) or intermediate domain processing. This method tested whether standard retrieval of text records is sufficient for complex clinical reasoning.
    \item \textbf{Single-agent RAG Baseline:} Executes metadata-based retrieval $R(p)$ and extracts structured facts to form the fact table $F$. However, instead of routing to specialized domain agents ($N_d$) or performing a final verification check, a single LLM agent jointly processes $q$, $C$, and $F$ in a single synthesis step to generate the response $\hat{y}$. This method tested whether fact extraction alone is sufficient without multi-agent domain specialization or verification.
\end{enumerate}

Table~\ref{tab:offline_updrs} and Table~\ref{tab:offline_combined} report the offline results across all four frameworks, for UPDRS-only and combined UPDRS and PDQ-8 queries. 
\begin{table}[ht]
\centering
\caption{Offline evaluation on \textbf{UPDRS-only queries}. Best results per column are in \textbf{bold}. $\uparrow$ indicates higher is better, $\downarrow$ indicates lower is better.}
\label{tab:offline_updrs}
\resizebox{0.85\linewidth}{!}{%
    \begin{tabular}{llcccc}
        \toprule
        \textbf{Task} & \textbf{Framework} & \textbf{Fact Precision $\uparrow$} & \textbf{Hallucination $\downarrow$} & \textbf{Temporal Fidelity $\uparrow$} & \textbf{SBERT Sim. $\uparrow$} \\
        \midrule
        \multirow{4}{*}{single-Visit}
            & Traditional          & 0.702 & 0.298 & 0.625 & 0.604 \\
            & RAG-only (baseline)  & 0.799 & 0.201 & 0.617 & \textbf{0.612} \\
            & Single-agent RAG     & 0.860 & 0.140 & 0.620 & 0.600 \\
            & MA-RAG (our)    & \textbf{0.940} & \textbf{0.060} & \textbf{0.641} & 0.610 \\
        \midrule
        \multirow{4}{*}{Longitudinal}
            & Traditional          & 0.436 & 0.564 & 0.630 & 0.597 \\
            & RAG-only (baseline)  & 0.652 & 0.348 & 0.581 & 0.661 \\
            & Single-agent RAG     & 0.780 & 0.220 & 0.610 & 0.623 \\
            & MA-RAG (our)   & \textbf{0.970} & \textbf{0.020} & \textbf{0.660} & \textbf{0.683} \\
        \bottomrule
    \end{tabular}
    }
\end{table}
\begin{table}[ht]
\centering
\caption{Offline evaluation on \textbf{combined UPDRS and PDQ-8 queries}. Best results per column are in \textbf{bold}. $\uparrow$ indicates higher is better, $\downarrow$ indicates lower is better.}
\label{tab:offline_combined}
\resizebox{0.85\linewidth}{!}{%
    \begin{tabular}{llcccc}
        \toprule
        \textbf{Task} & \textbf{Framework} & \textbf{Fact Precision $\uparrow$} & \textbf{Hallucination $\downarrow$} & \textbf{Temporal Fidelity $\uparrow$} & \textbf{SBERT Sim. $\uparrow$} \\
        \midrule
        \multirow{4}{*}{single-Visit}
            & Traditional          & 0.639 & 0.361 & 0.623 & \textbf{0.634} \\
            & RAG-only (baseline)  & 0.708 & 0.292 & 0.576 & 0.585 \\
            & Single-agent RAG     & 0.820 & 0.180 & \textbf{0.700} & 0.620 \\
            & MA-RAG (our)   & \textbf{0.930} & \textbf{0.070} & 0.643 & 0.633 \\
        \midrule
        \multirow{4}{*}{Longitudinal}
            & Traditional          & 0.592 & 0.408 & 0.605 & 0.679 \\
            & RAG-only (baseline)  & 0.652 & 0.348 & 0.581 & 0.661 \\
            & Single-agent RAG     & 0.765 & 0.291 & 0.600 & 0.590 \\
            & MA-RAG (our)    & \textbf{0.990} & \textbf{0.010} & \textbf{0.620} & \textbf{0.722} \\
        \bottomrule
    \end{tabular}
    }
\end{table}

MA-RAG consistently achieves the highest Fact Precision and lowest Hallucination Rate across both single-visit (single-session) and longitudinal tasks. The performance gains are most pronounced in longitudinal queries, where reasoning complexity is highest: Fact Precision increases from $0.436$ (Traditional) to $0.652$ (RAG-only), $0.780$ (Single-agent RAG), and reaches $0.970$ in MA-RAG for UPDRS-only queries, with a similar trajectory observed in the combined setting ($0.592$, $0.652$, $0.765$, and $0.990$, respectively). These improvements directly validate our architectural design choices. First, introducing metadata-based retrieval (RAG-only) reduces reliance on static context window parsing. Second, extracting structured fact tables $F$ (Single-agent RAG) significantly lowers hallucinations compared to reasoning over raw unstructured text passages. Finally, isolating domain processing into specialized agents and executing final verification (MA-RAG) provides the critical final jump, effectively preserving factual correctness and eliminating cross-domain interference as clinical context scales.

In contrast, surface-level evaluation metrics such as SBERT Similarity and Temporal Fidelity do not reliably reflect factual correctness. For instance, in the single-visit UPDRS+PDQ-8 setting, Single-agent RAG achieves the highest Temporal Fidelity ($0.700$) despite exhibiting a noticeably higher hallucination rate ($0.180$) than MA-RAG ($0.070$). Similarly, Traditional RAG achieves the highest SBERT similarity ($0.634$) in that same setting despite lower fact precision ($0.639$). This demonstrates that semantic similarity and temporal ordering alone are insufficient for evaluating clinical fidelity, reinforcing the necessity of strict fact-based evaluation metrics.

To validate these observations, we performed paired $t$-tests~\cite{tTest1908} using query-level scores to compare MA-RAG against each baseline across all tasks and metrics. All comparisons were statistically significant ($p < 0.05$), confirming that the superior factual precision of MA-RAG is robust and consistent.

\subsubsection{Comparison Across Different LLMs}
In a second offline experiment, we assessed how MA-RAG’s performance varies when the underlying LLM is replaced, while keeping the multi-agent architecture, retrieval pipeline, and evaluation protocol fixed. To promote reproducibility and facilitate adoption by the broader research community, we restrict the comparison to open-source LLMs with up to 8B parameters while balancing reasoning capability, inference latency, and computational cost.  Specifically, we repeated the longitudinal and single-visit UPDRS and UPDRS+PDQ-8 tasks from the previous experiment using four open-source backbones (Phi-2~\cite{Javaheripi2023Phi2}, OpenBioLLM-8B~\cite{Chen2025}, Meditron-7B~\cite{chen2023meditron}, and Llama-3.1-8B-Instruct). Then, we measured Fact Precision, Hallucination, Temporal Fidelity, and SBERT Similarity under each configuration. The results for longitudinal queries are summarized in Table~\ref{tab:offline_llm_longitudinal}, and the single-visit results in Table~\ref{tab:offline_llm_pervisit}.
\begin{table}[ht]
\centering
\caption{Offline evaluation of MA-RAG with different language models on \textbf{longitudinal queries}. Best results per column are in \textbf{bold}.}
\label{tab:offline_llm_longitudinal}
\resizebox{0.85\linewidth}{!}{%
\begin{tabular}{llcccc}
\toprule
\textbf{Task} & \textbf{LLM} & \textbf{Fact Precision} & \textbf{Hallucination} & \textbf{Temporal Fidelity} & \textbf{SBERT Sim.} \\
\midrule
\multirow{6}{*}{UPDRS}
  & Phi-2            & 0.650 & 0.350 & 0.867 & 0.636 \\
  & OpenBioLLM-8B       & 0.725 & 0.275 & \textbf{1.000} & \textbf{0.699} \\
  & Meditron-7B         & 0.513 & 0.487 & \textbf{1.000} & 0.341 \\
  & LLaMA-3.1-8B-Instruct & \textbf{0.970} & \textbf{0.020} & 0.660 & 0.683 \\
\midrule
\multirow{6}{*}{UPDRS + PDQ-8}
  & Phi-2            & 0.730 & 0.270 & \textbf{1.000} & 0.608 \\
  & OpenBioLLM-8B       & 0.920 & 0.080 & 0.857 & \textbf{0.726} \\
  & Meditron-7B         & 0.751 & 0.249 & \textbf{1.000} & 0.488 \\
  & LLaMA-3.1-8B-Instruct & \textbf{0.990} & \textbf{0.010} & 0.620 & 0.722 \\
\bottomrule
\end{tabular}
}
\end{table}
\begin{table}[ht]
\centering
\caption{Offline evaluation of MA-RAG with different language models on \textbf{single-visit queries}. Best results per column are in \textbf{bold}.}
\label{tab:offline_llm_pervisit}
\resizebox{0.85\linewidth}{!}{%
\begin{tabular}{llcccc}
\toprule
\textbf{Task} & \textbf{LLM} & \textbf{Fact Precision} & \textbf{Hallucination} & \textbf{Temporal Fidelity} & \textbf{SBERT Sim.} \\
\midrule
\multirow{6}{*}{UPDRS}
  & Phi-2            & 0.759 & 0.241 & 0.667 & 0.588 \\
  & OpenBioLLM-8B       & 0.922 & 0.078 & 0.897 & 0.477 \\
  & Meditron-7B         & 0.745 & 0.255 & \textbf{1.000} & 0.420 \\
  & LLaMA-3.1-8B-Instruct & \textbf{0.940} & \textbf{0.060} & 0.641 & \textbf{0.610} \\
\midrule
\multirow{6}{*}{UPDRS + PDQ-8}
  & Phi-2            & 0.829 & 0.171 & 0.667 & 0.595 \\
  & OpenBioLLM-8B       & 0.929 & 0.071 & 0.750 & 0.441 \\
  & Meditron-7B         & 0.699 & 0.301 & \textbf{1.000} & 0.436 \\
  & LLaMA-3.1-8B-Instruct & \textbf{0.930} & \textbf{0.070} & 0.643 & \textbf{0.633} \\
\bottomrule
\end{tabular}
}
\end{table}

Across all evaluation tasks, Llama-3.1-8B-Instruct consistently achieves the highest Fact Precision and lowest Hallucination rate. For longitudinal queries (Table~\ref{tab:offline_llm_longitudinal}), it reaches a Fact Precision of $0.970$ (UPDRS) and $0.990$ (UPDRS+PDQ-8) with hallucination rates as low as $0.020$ and $0.010$, substantially outperforming specialized biomedical models such as OpenBioLLM-8B and Meditron-7B. A similar trend is observed in single-visit queries (Table~\ref{tab:offline_llm_pervisit}), where Llama-3.1-8B-Instruct maintains lead precision scores ($0.940$ and $0.930$).

Interestingly, while Meditron-7B and OpenBioLLM-8B attain perfect or near-perfect Temporal Fidelity (up to $1.000$), this does not correspond to factual accuracy. Models like Meditron-7B strictly preserve visit chronology but frequently hallucinate or omit specific numerical domain scores, resulting in lower Fact Precision (e.g., $0.513$ in longitudinal UPDRS). Conversely, generalist instruction-tuned models like Llama-3.1-8B-Instruct possess superior instruction-following capabilities required to strictly adhere to the structured fact table $F$ across specialized agent prompts. This further reinforces that semantic similarity and temporal ordering alone are insufficient metrics for verifying clinical factuality.

\subsection{Subjective Evaluation}
For the subjective evaluation, we deployed MA-RAG as a Gradio application. Since the system is not a conversational tool, we generated a fixed set of eight representative clinical summaries, comprising two summaries from each of the four supported query categories: single-session, longitudinal, comparison, and cohort. Three certified clinicians independently evaluated the summaries through Google Forms using a five-point Likert scale, ranging from 1 (strongly disagree) to 5 (strongly agree), in response to the following six questions: 1) the clinical information in this summary is factually correct, 2) all key clinical findings relevant to the query are captured, 3) the summary is focused and directly answers the query, 4) the summary is well-organized and easy to understand, 5) I did not detect any fabricated, misleading, or unsafe information, and 6) I would be comfortable acting on or sharing this summary clinically. We additionally asked a seventh open-ended question: whether the clinician noticed a specific problem and, if so, to describe it briefly. We report the average rating for each query category across questions 1 to 6, and retain the open-ended responses for future improvements.

\begin{figure}
  \centering
  \includegraphics[width=0.8\linewidth]{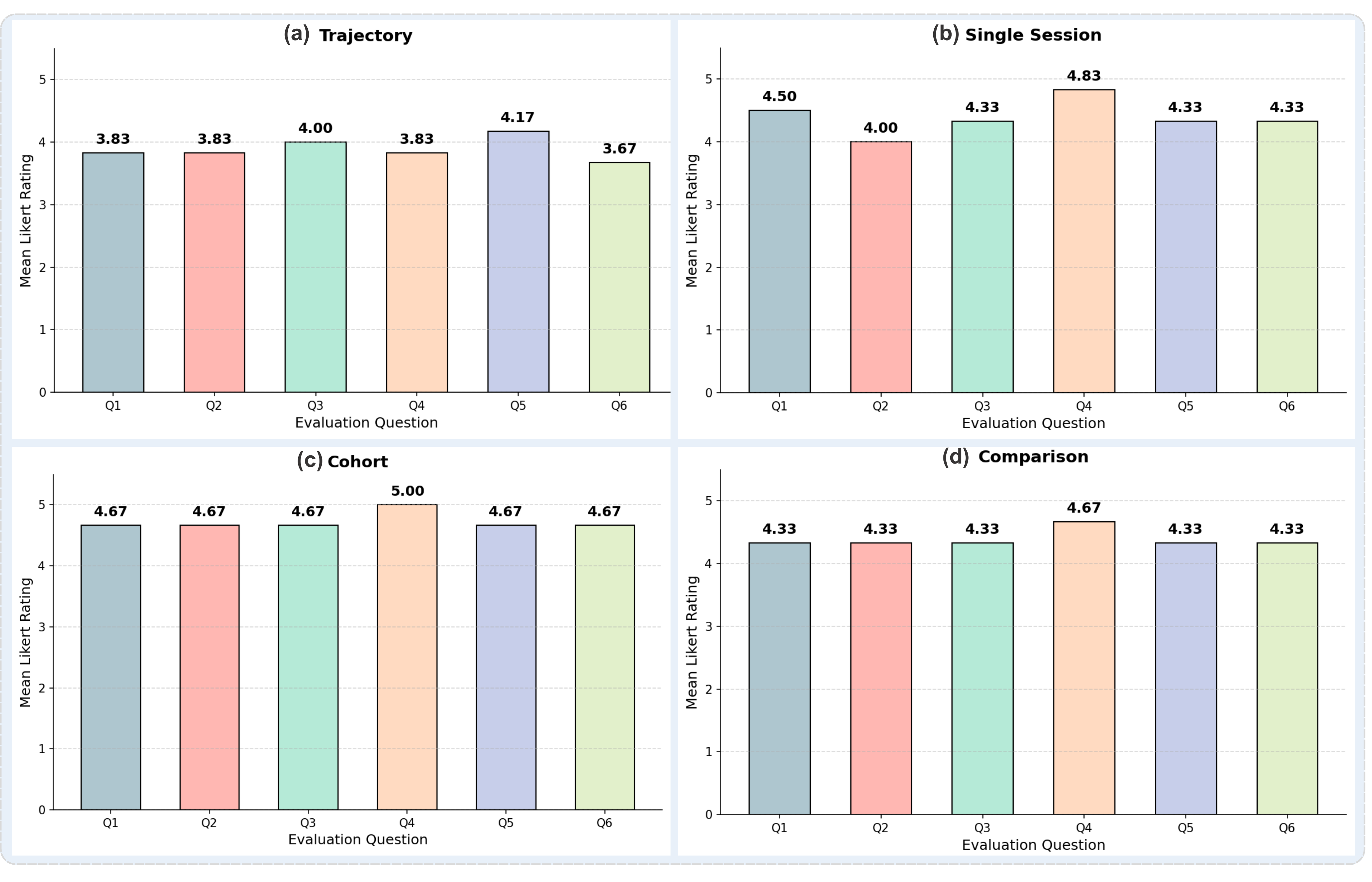}
  \caption{Subjective evaluation results: Mean clinician Likert ratings (n=3) across Q1--Q6 for (a) Trajectory, (b) Single Session, (c) Cohort, and (d) Comparison query summaries.}
  \label{fig:subjective_eval}
\end{figure}
Figure~\ref{fig:subjective_eval} presents the mean clinician ratings (n=3) across the six Likert questions for each of the four query categories. Overall, clinicians rated all summary types positively, with mean scores ranging from 3.67 to 5.00. Cohort summaries received the highest and most consistent ratings (4.67--5.00), reflecting strong agreement on clarity and usefulness for population-level analysis. Single Session and Comparison summaries were also rated highly, with organization (Q4) receiving the highest score for Single Session summaries (4.83). Trajectory summaries received comparatively lower ratings (3.67--4.17), particularly for clinical actionability (Q6), suggesting that longitudinal summaries remain the most challenging task. Detailed outputs for each category are provided in Appendix~\ref{app:ex_outputs}.

Clinicians provided additional context for these ratings in open-ended feedback (Q7), where they consistently praised the factual content and organization of the summaries, particularly for Cohort and Comparison queries. For Trajectory summaries, clinicians identified several opportunities for improvement, with suggestions of a) incorporating intervention or treatment context to better explain changes across visits, b) refining cross-domain reasoning in a small number of cases involving discordance or longitudinal interpretation, and c) using clearer clinical terminology and more concise language to support the clinical read-interpret-decision workflow better. These findings suggest that MA-RAG produces clinically useful summaries across all query types while highlighting directions for further improving longitudinal clinical summarization.

\subsection{Computational Cost, Ethical Considerations and Limitations}
Since execution time depends on hardware and deployment settings, we estimated the computational cost by the number of LLM inference calls required for each analysis type. As shown in Table~\ref{tab:computational_cost}, Trajectory and Single-Session analyses invoke six agents (Planner, four domain-specific agents, and Final), corresponding to a relative cost of 6.0$\times$ a single-pass RAG baseline. In contrast, Comparison and Cohort analyses require only three LLM calls (Planner, one specialized agent, and Final), resulting in a relative cost of 3.0$\times$. This increase in inference cost reflects the additional coordination required for domain-specialized reasoning, particularly for longitudinal analyses involving multiple clinical domains and visits. Despite the higher computational cost, the objective and subjective evaluations demonstrate that this design substantially improves factual accuracy while reducing hallucinations compared with single-pass baselines.
\begin{table}[h]
\centering
\caption{Computational Cost by Analysis Type (LLM Inference-Based)}
\label{tab:computational_cost}
\resizebox{0.85\linewidth}{!}{%
    \begin{tabular}{lccc}
        \hline
        \textbf{Query Type} & \textbf{LLM Agents Invoked} & \textbf{\# LLM Calls} & \textbf{\makecell[c]{Relative Cost\\ (× RAG baseline)}} \\
        \hline
        Trajectory / Single-Session & \makecell[c]{Planner + \\Motor + ADL + Non-motor + QoL +\\ Final} & 6 & 6.0× \\ \hline
        Comparison & Planner + Comparison Agent + Final & 3 & 3.0× \\ \hline
        Cohort & Planner + Cohort Agent + Final & 3 & 3.0× \\
        \hline
    \end{tabular}
    }
\end{table}

From an ethical perspective, the additional computation is justified because it is allocated to the most clinically demanding tasks, where factual errors can directly affect clinical interpretation and decision-making. The system further prioritizes patient privacy by deploying an open-source LLM entirely on local infrastructure, ensuring that sensitive clinical data never leaves the institutional environment. In addition, the interface restricts users to predefined query templates rather than unrestricted free-text input, reducing the risk of prompt injection and other adversarial interactions while improving the consistency of generated summaries.

The current implementation of MA-RAG is instantiated for Parkinson's disease assessments using the UPDRS and PDQ-8 instruments. While the domain-specific reasoning agents, prompts, and extraction rules are tailored to these assessments, the overall architecture is modular. Extending the framework to another disease primarily requires adapting the domain-specific reasoning layer. Future work will evaluate MA-RAG on multimodal assessment datasets from additional clinical domains.

\section{Conclusion ad Future Work}\label{sect:conclusion}
This paper presented MA-RAG, a multi-agent retrieval-augmented generation framework for generating longitudinal multimodal clinical summaries of Parkinson's disease from UPDRS and PDQ-8 assessments. By combining domain-specialized agents, structured fact extraction, and a final verification stage, MA-RAG produces clinically coherent summaries while maintaining strong factual grounding. In offline evaluations, MA-RAG consistently outperformed Traditional, RAG-only, and Single-agent RAG baselines, achieving up to a 122\% relative improvement in Fact Precision (0.436 to 0.990) and reducing Hallucination rates by up to 98\% (0.564 to 0.010) compared to the Traditional baseline. Subjective evaluation by clinical experts further demonstrated that the generated summaries were clinically meaningful and well organized across all four analysis types, supporting the potential of MA-RAG as a trustworthy clinical decision-support tool for longitudinal patient assessment.

Future work will focus on extending MA-RAG beyond Parkinson's disease by incorporating additional neurological disorders and heterogeneous clinical data sources, including electronic health records, imaging reports, laboratory results, and wearable sensor data. We also plan to incorporate clinician feedback through memory-augmented adaptation to improve longitudinal reasoning without fine-tuning the underlying LLM~\cite{zhou2025Memento}, and to validate the framework in larger prospective clinical studies involving more clinicians, institutions, and real-world deployment settings.

\section*{Resource Availability}
To facilitate reproducibility and further research, we make our full open-source codebase, including de-identified Knowledge Base datasets, and an interactive Gradio interface publicly available at: \url{https://github.com/txst-cs-smartfall/clinical-LLM-MA-RAG}.

\section*{Acknowledgment}
This research was funded by the National Science Foundation (NSF) under the Smart and Connected Health (SCH) Program grant number 21223749.

\clearpage
\bibliographystyle{model1-num-names}

\bibliography{cas-refs}

\clearpage
\appendix
\section{Appendix}
\section{Static Clinical Context}\label{app:static_contexts}

Figure~\ref{fig:static_contexts} shows static domain contexts $C_d$ for $d \in \{\text{motor}, \text{ADL}, \text{nonmotor}, \text{QoL}\}$, each encoding scale definitions, item-level scoring rules, clinical factor domains, and reliability statistics for its respective domain agent. The combined static context $C = \bigcup_d C_d$ is distributed to the corresponding agents in place of Knowledge Base retrieval, ensuring complete and consistent domain knowledge at inference time.
\begin{center}
  \includegraphics[width=\linewidth]{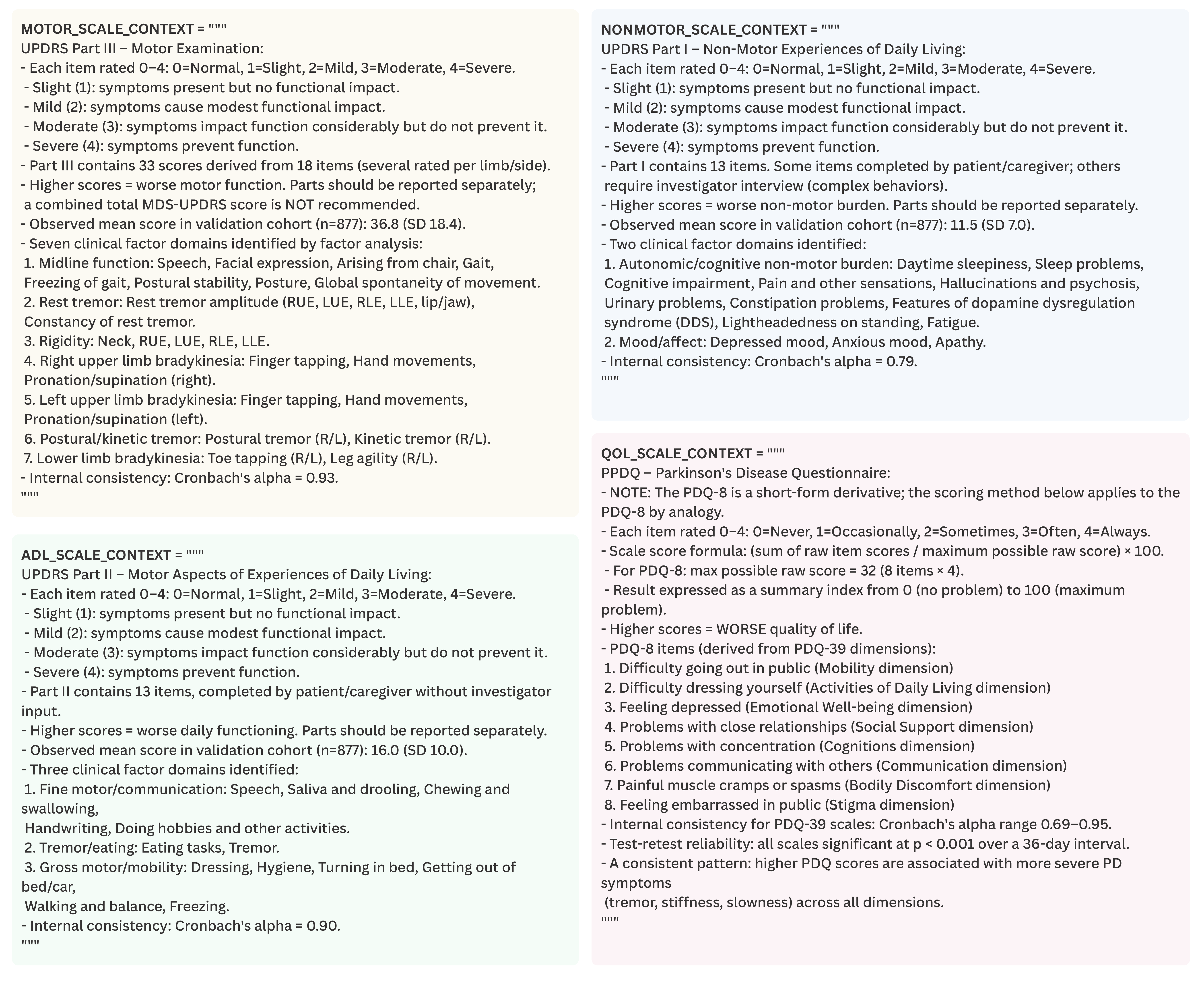}
  \captionof{figure}{Static domain contexts}
  \label{fig:static_contexts}
\end{center}

\section{System Prompts}\label{app:system_prompts}
\begin{tcolorbox}[
    breakable,
    enhanced,
    colback=LightGreen,
    colframe=PromptGreen,
    colbacktitle=PromptGreen,
    coltitle=white,
    title={System Prompt For Orchestrator (\texttt{plan\_node})},
    fonttitle=\bfseries\small,
]

\begin{lstlisting}
You are a clinical query planner for a Parkinson disease analysis system.
Convert the user's question into a structured analysis plan.

Rules:
1. PARTICIPANTS must be patient IDs mentioned in the query.
2. If the question refers to all patients use PARTICIPANTS: ALL.
3. ANALYSIS_TYPE must be one of: trajectory, comparison, cohort, single-session.
4. METRICS must be one or both of: UPDRS, PDQ8.
5. FILTER must be one of:
   none | baseline | latest_visit | baseline_vs_latest | first_N_visits | last_N_visits
   where N is the exact number specified by the user (e.g. first_2_visits, last_5_visits).

ANALYSIS_TYPE Selection Rules:
- Use 'comparison'     when the query contains: compare, vs, versus, difference between, contrast, side by side
- Use 'trajectory'     when the query contains: trajectory, progression, over time, trend, how did, changed, visits
- Use 'cohort'         when the query contains: highest, lowest, average, fastest, most, least, all patients, who has
- Use 'single-session' when the query contains: baseline, latest, last visit, most recent, session

FILTER Selection Rules:
- Use 'none'               for trajectory or comparison queries (all visits needed)
- Use 'baseline'           when the query asks about the first/baseline visit only
- Use 'latest_visit'       when the query asks about the most recent/last visit only
- Use 'baseline_vs_latest' when the query asks to compare first vs last visit
- Use 'first_N_visits'     when the query asks about the first/earliest N visits - replace N with the user's number
- Use 'last_N_visits'      when the query asks about the last/most recent N visits - replace N with the user's number

METRICS Selection Rules:
- Use 'UPDRS'        when motor, ADL, or non-motor domains are mentioned
- Use 'PDQ8'         when quality of life, QoL, or PDQ8 is mentioned
- Use 'UPDRS, PDQ8'  when the query is general (no specific metric mentioned)

Examples:
Query: compare trajectories of proj ids 2 and 1
PARTICIPANTS: 2, 1
ANALYSIS_TYPE: comparison
METRICS: UPDRS, PDQ8
FILTER: none

Query: show me the motor progression of patient 3
PARTICIPANTS: 3
ANALYSIS_TYPE: trajectory
METRICS: UPDRS
FILTER: none

Query: who has the highest PDQ8 score
PARTICIPANTS: ALL
ANALYSIS_TYPE: cohort
METRICS: PDQ8
FILTER: none

Query: what is the baseline status of patient 5
PARTICIPANTS: 5
ANALYSIS_TYPE: single-session
METRICS: UPDRS, PDQ8
FILTER: baseline

Query: compare the latest visit of patients 1 and 4
PARTICIPANTS: 1, 4
ANALYSIS_TYPE: comparison
METRICS: UPDRS, PDQ8
FILTER: latest_visit

Query: show me the first 2 visits of patient 3
PARTICIPANTS: 3
ANALYSIS_TYPE: trajectory
METRICS: UPDRS, PDQ8
FILTER: first_2_visits

Query: compare the last 4 visits of patients 1 and 2
PARTICIPANTS: 1, 2
ANALYSIS_TYPE: comparison
METRICS: UPDRS, PDQ8
FILTER: last_4_visits

Query: show trajectory for patient 7 over the last 5 visits
PARTICIPANTS: 7
ANALYSIS_TYPE: trajectory
METRICS: UPDRS, PDQ8
FILTER: last_5_visits

Query: what is the status of visit 5 for patient 1
PARTICIPANTS: 1
ANALYSIS_TYPE: single-session
METRICS: UPDRS, PDQ8
FILTER: visit_5

Query: what is the status of visit 2 for patient 3
PARTICIPANTS: 3
ANALYSIS_TYPE: single-session
METRICS: UPDRS, PDQ8
FILTER: visit_2

Return ONLY the plan using this exact format:
PARTICIPANTS: ...
ANALYSIS_TYPE: ...
METRICS: ...
FILTER: ...
\end{lstlisting}
\end{tcolorbox}

\section{Example Outputs}\label{app:ex_outputs}

\begin{tcolorbox}[
    breakable,
    enhanced,
    colback=LightGreen,
    colframe=PromptGreen,
    colbacktitle=PromptGreen,
    coltitle=white,
    title={Query Type: Trajectory | Query: Show me the UPDRS trajectory of proj id 1},
    fonttitle=\bfseries\small,
]

\begin{lstlisting}
==============================
FINAL CLINICAL SUMMARY
==============================
Patient 1's Motor Trajectory
At the baseline visit on January 17, 2020, the patient had a Motor score of 14, indicating mild motor symptoms. By March 7, 2020, the Motor score worsened to 18, a +4 increase from the baseline and a +4 increase from the previous visit. This worsening was driven by increases in several individual items, including speech impairment, facial expression, and bradykinesia. However, by April 28, 2020, the Motor score improved significantly to 7, a -11 decrease from the previous visit and a -7 decrease from the baseline. This improvement was largely due to decreases in rigidity, finger tapping, and rapid alternating movements of hands. Unfortunately, by June 14, 2020, the Motor score worsened again to 20, a +13 increase from the previous visit and a +6 increase from the baseline. This worsening was driven by increases in rest tremor, postural tremor, and bradykinesia. Finally, by October 5, 2020, the Motor score improved dramatically to 0, a -20 decrease from the previous visit and a -14 decrease from the baseline. This near-complete resolution of motor symptoms was likely due to the resolution of all individual items.

Patient 1's ADL Trajectory
At the baseline visit on January 17, 2020, the patient had an ADL score of 13, indicating some level of impairment in fine motor skills and communication. By March 7, 2020, the ADL score worsened to 22, a +9 increase from the baseline and a +9 increase from the previous visit. This worsening was driven by increases in eating difficulty, freezing of gait, and handwriting difficulty. However, by April 28, 2020, the ADL score improved significantly to 3, a -10 decrease from the baseline and a -19 decrease from the previous visit. This improvement was largely due to decreases in turning in bed difficulty and falls. Unfortunately, by June 14, 2020, the ADL score worsened again to 23, a +20 increase from the previous visit and a +20 increase from the baseline. This worsening was driven by increases in speech difficulty, handwriting difficulty, eating difficulty, and freezing of gait. Finally, by October 5, 2020, the ADL score improved dramatically to 0, a -13 decrease from the baseline and a -17 decrease from the previous visit. This near-complete resolution of ADL difficulties was likely due to the resolution of all individual items.

Patient 1's Non-Motor Trajectory
At the baseline visit on January 17, 2020, the patient had a NonMotor score of 13, indicating a moderate non-motor burden. By March 7, 2020, the NonMotor score increased to 15, a +2 increase from the previous visit and a +2 increase from the baseline. This increase was driven by worsening sleep problems. However, by April 28, 2020, the NonMotor score decreased to 7, a -8 decrease from the previous visit and a -6 decrease from the baseline. This decrease was largely due to an increase in anxiety. Unfortunately, by June 14, 2020, the NonMotor score worsened again to 19, a +12 increase from the previous visit and a +6 increase from the baseline. This worsening was driven by the emergence of hallucinations and psychosis. Finally, by October 5, 2020, the NonMotor score improved dramatically to 0, a -19 decrease from the previous visit and a -13 decrease from the baseline. This near-complete resolution of non-motor symptoms was likely due to the resolution of all individual items.

Cross-Domain Analysis
Discordance: At Visit 4, the Motor score worsened, while the ADL score worsened, and the NonMotor score worsened. This discordance suggests that the patient's motor symptoms were not improving despite improvements in ADLs and non-motor symptoms.
Leading Indicators: The NonMotor score worsened first, indicating that the patient's non-motor symptoms were the earliest warning signs of disease progression.
Lagging Indicators: The ADL score took the longest to improve, indicating that the patient's daily functioning was the most challenging aspect of their disease to recover from.
Floor Effects: At Visit 5, the Motor, ADL, and NonMotor scores all reached 0, indicating complete resolution of symptoms. However, it is unlikely that these symptoms completely resolved, as the patient still reported some QoL issues.

Domains Requiring Monitoring
Based on the trajectory patterns observed, the Motor domain requires closest clinical monitoring, particularly the bradykinesia item, which worsened at Visit 4 and did not fully resolve at Visit 5. Additionally, the ADL domain requires monitoring, as the patient's daily functioning was severely impaired at Visit 4 and only partially recovered at Visit 5.
\end{lstlisting}
\end{tcolorbox}
\begin{tcolorbox}[
    breakable,
    enhanced,
    colback=LightGreen,
    colframe=PromptGreen,
    colbacktitle=PromptGreen,
    coltitle=white,
    title={Query Type: Single Session | Query: What is the baseline status of proj id 1?},
    fonttitle=\bfseries\small,
]

\begin{lstlisting}
==============================
FINAL CLINICAL SUMMARY
==============================
Here is the concise final clinical summary:
Visit 1 (2020-01-17)
Motor: Total score = 14. Notable items include:
Speech impairment (motor exam) = 2
Facial expression (hypomimia) = 2
Posture = 2
ADL: Total score = 13. Significant difficulties with:
Speech difficulty = 2
Handwashing difficulty = 1
Salivation problems = 2
Non-Motor: Total score = 13. Primary concerns:
Sleep problems = 3
Daytime sleepiness = 2
PDQ-8: Total score = 6. High-impact symptoms:
Difficulty going out in public
Difficulty dressing
Problems with concentration

Clinical Interpretation: This patient presents with a complex profile of motor, ADL, and non-motor symptoms at baseline, indicating a need for comprehensive management to address both physical and psychological aspects of Parkinson's disease.

Cross-Domain Patterns: There is no discordance between domains. However, the Non-Motor domain carries a disproportionate burden compared to the other domains, suggesting that the patient's current disease expression is heavily weighted towards autonomic and cognitive issues.

Domains Requiring Closest Follow-Up: The Non-Motor domain requires closest follow-up due to its high score and the presence of significant sleep problems and daytime sleepiness. Additionally, the ADL domain warrants attention given the patient's difficulties with speech, handwashing, and salivation problems, which may impact daily functioning and independence.
\end{lstlisting}
\end{tcolorbox}
\begin{tcolorbox}[
    breakable,
    enhanced,
    colback=LightGreen,
    colframe=PromptGreen,
    colbacktitle=PromptGreen,
    coltitle=white,
    title={Query Type: Cohort | Query: Which proj id is at high risk of disease progression?},
    fonttitle=\bfseries\small,
]

\begin{lstlisting}
==============================
FINAL CLINICAL SUMMARY
==============================
Patient with Project ID 20 is at high risk of disease progression due to significant worsening of motor symptoms (largest motor increase of 16 points) and rapid decline in quality of life (fastest QOL worsening slope of 8.0).
\end{lstlisting}
\end{tcolorbox}
\begin{tcolorbox}[
    breakable,
    enhanced,
    colback=LightGreen,
    colframe=PromptGreen,
    colbacktitle=PromptGreen,
    coltitle=white,
    title={Query Type: Comparison | Query: Compare the disease progression between proj id 4 and proj id 7.},
    fonttitle=\bfseries\small,
]

\begin{lstlisting}
==============================
FINAL CLINICAL SUMMARY
==============================
Cross-Domain Pattern Analysis

Upon examining the cross-domain patterns, we notice several key observations:
DISCORDANCE: One notable discordance is observed in the ADL domain, where Patient 4 experiences a significant improvement (-9 points) followed by a worsening, whereas Patient 7's ADL score worsens gradually without such a dramatic reversal. This suggests that Patient 4 may be experiencing fluctuations in their ability to perform daily activities, potentially due to changes in medication or environmental factors.
DIVERGENCE POINT: The divergence point occurs between Visit 5 and Visit 6 for Patient 4, where their motor score improves (-3 points), while their ADL score worsens (+4 points). This indicates a decoupling between motor function and daily living abilities, highlighting the complexity of Parkinson's disease progression.
COMPENSATORY PATTERNS: Interestingly, despite having worse motor symptoms, Patient 4 exhibits a better quality of life (lower PDQ8 score) than Patient 7 at certain visits (e.g., Visit 5). This might suggest that Patient 4 has developed compensatory strategies to cope with their motor symptoms, maintaining a relatively better quality of life.

Domains Requiring Monitoring
For each patient, the following domains require closest monitoring:
Patient 4: The ADL domain requires closest monitoring, particularly after Visit 5, as it shows a significant worsening (-9 points) followed by an improvement, indicating potential fluctuations in daily functioning.
Patient 7: The PDQ8 domain requires closest monitoring, as it consistently shows a worsening trend, indicating a decrease in quality of life over time.

Additional Observations
Shared Disease Milestones: There are no clear synchronized worsening periods across patients suggesting shared disease milestones.
Domain Divergence: While Patient 4's motor symptoms worsen, their ADL score improves initially before worsening again, indicating a complex relationship between motor function and daily living abilities.
Mismatch between UPDRS Burden and PDQ8 Quality of Life: Patient 4 exhibits a mismatch between their motor symptoms (worsening) and quality of life (better PDQ8 score at certain visits), suggesting effective coping mechanisms or treatment responses.
Clinical Inflection Points: Visits 5 and 6 for Patient 4 represent potential clinical inflection points, where multiple domains show significant changes (e.g., ADL worsening, motor improvement).
Faster Disease Velocity: Patient 4 appears to have a faster overall disease velocity, with significant fluctuations in motor, ADL, and non-motor symptoms throughout the study period.
\end{lstlisting}
\end{tcolorbox}

% ...


\end{document}